\documentclass[lettersize,journal]{IEEEtran}
\usepackage{amsmath,amsfonts}
\usepackage{algorithmic}
\usepackage{algorithm}
\usepackage{array}
\usepackage[caption=false,font=normalsize,labelfont=sf,textfont=sf]{subfig}
\usepackage{textcomp}
\usepackage{stfloats}
\usepackage{url}
\usepackage{verbatim}
\usepackage{graphicx}
\usepackage{cite}
\usepackage{multirow}
\usepackage{capt-of}
\usepackage{pifont}
\usepackage{graphicx}
\usepackage{capt-of}
\usepackage{cuted}

\begin{document}

\title{ALPHA-$\alpha$ and Bi-ACT Are All You Need: Importance of Position and Force Information/Control for Imitation Learning of Unimanual and Bimanual Robotic Manipulation with Low-Cost System}

\author{
\IEEEauthorblockN{Masato Kobayashi$^{1,2,3*}$, Thanpimon Buamanee$^{2}$, Takumi Kobayashi$^{3}$}
\thanks{
$1$ D3 Center, Osaka University, Japan,
$2$ Graduate School of Information Science and Technology of Osaka University, Japan,
$3$ School of Engineering Science, Osaka University, Japan,
$^{*}$Corresponding author
}
}

\maketitle

\begin{strip}
  \vspace{-75pt}
\centering
\includegraphics[width=\textwidth]{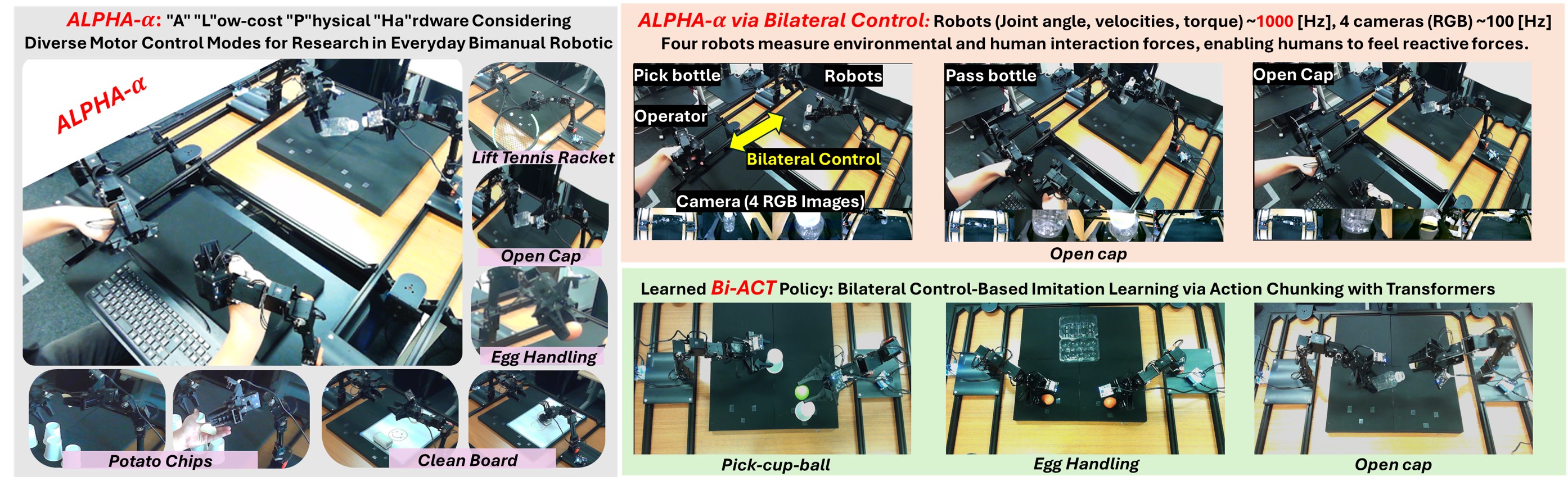}
\captionof{figure}{Overview of Bilateral Control-Based Imitation Learning System using ALPHA-$\alpha$ and Bi-ACT}
\label{fig:alpha-biact}
\end{strip}

\begin{abstract}
Autonomous manipulation in everyday tasks requires flexible action generation to handle complex, diverse real-world environments, such as objects with varying hardness and softness. Imitation Learning (IL) enables robots to learn complex tasks from expert demonstrations. However, a lot of existing methods rely on position/unilateral control, leaving challenges in tasks that require force information/control, like carefully grasping fragile or varying-hardness objects. As the need for diverse controls increases, there are demand for low-cost bimanual robots that consider various motor inputs.
To address these challenges, we introduce Bilateral Control-Based Imitation Learning via Action Chunking with Transformers(Bi-ACT) and"A" "L"ow-cost "P"hysical "Ha"rdware Considering Diverse Motor Control Modes for Research in Everyday Bimanual Robotic Manipulation (ALPHA-$\alpha$).
Bi-ACT leverages bilateral control to utilize both position and force information, enhancing the robot's adaptability to object characteristics such as hardness, shape, and weight. The concept of ALPHA-$\alpha$ is affordability, ease of use, repairability, ease of assembly, and diverse control modes (position, velocity, torque), allowing researchers/developers to freely build control systems using ALPHA-$\alpha$.
In our experiments, we conducted a detailed analysis of Bi-ACT in unimanual manipulation tasks, confirming its superior performance and adaptability compared to Bi-ACT without force control. Based on these results, we applied Bi-ACT to bimanual manipulation tasks using ALPHA-$\alpha$. Experimental results demonstrated high success rates in coordinated bimanual operations across multiple tasks, verifying the effectiveness of our approach in complex real-world scenarios.
The effectiveness of the Bi-ACT and ALPHA-$\alpha$ can be seen through comprehensive real-world experiments.
Video available at: https://mertcookimg.github.io/alpha-biact/
\end{abstract}

\section{Introduction}
Autonomous manipulation in real-world tasks remains a considerable challenge due to varied environments that necessitate the generation of adaptive motions.
Recent advancements in robotic manipulation have been driven by Imitation Learning (IL), which enables robots to learn complex tasks from expert demonstrations\cite{biact2024pan, IMI2024IA1, IMI2024IA2, IMI2024IA3}. These approaches have been instrumental in enhancing the adaptability and proficiency of robots in handling complex manipulation tasks.

Critical to IL's success is effective data collection\cite{lfd2009argall, lfd2022mukherjee}, often facilitated by teleoperated systems such as virtual reality headsets with hand-tracking mechanisms\cite{tre2018Zhang}, smartphones\cite{tre2021tung}, keyboard inputs\cite{tre2018fan}, UMI\cite{chi2024universal}, and leader-follower systems\cite{ge2023swu, act2023zhao}.
In particular, systems like ALOHA\cite{act2023zhao,team2024aloha,zhao2024aloha} and Mobile ALOHA\cite{mact2024fu}, being low-cost, facilitate a broader participation of researchers and developers in building and experimenting with these systems, thereby significantly advancing the field of research. Furthermore, the availability of an affordable system for data collection is essential for robotics research, such as the development of robotic foundational models.

ALOHA\cite{act2023zhao,team2024aloha,zhao2024aloha}, and Mobile ALOHA\cite{mact2024fu} have shown significant advancements in data collection methods by using leader-follower systems that collect robot joint angles along with image data by unilateral control.
Unilateral control is a method in which control commands are issued in one direction from the operator to the robot, with no sensory feedback loop to adjust actions based on interaction with the environment.
However, the lack of force information and feedback control presents a limitation, making it difficult to grasp object characteristics such as hardness, shape, and weight.
Therefore there is growing interest in data collection through bilateral control systems that can process both position and force information/control without the use of force/tactile sensors\cite{biact2024pan, IMIB2022sakaino}.

In the field of IL, model selection is crucial for robots to accurately understand and replicate complex behaviors. Traditional methods, such as RNN, Long Short-Term Memory (LSTM)\cite{lstm} and Transformer\cite{TRANS2017vaswani}, have been used to process time-series data but suffer from a 'compounding error' issue, where small mistakes in action prediction lead to larger errors over time\cite{con2021ke}. Action Chunking with Transformers (ACT)\cite{act2023zhao}, trained with data collected via ALOHA, has overcome these challenges by predicting multiple future steps, thereby reducing error accumulation.  This approach also addresses issues like pauses during demonstrations, which are difficult to model using Markovian single-step policies\cite{con2022swamy}.

Addressing current limitations in IL, Action Chunking with Transformers(ACT) method learned from data on robot joint angles and images, without incorporating force information and control\cite{act2023zhao, mact2024fu}.
In contrast, bilateral control-based imitation learning\cite{IMIB2022sakaino} utilizes both position and force information/control, but most methods use LSTM as model\cite{lstm}, limiting complex task performance.
In addition, there are methods using Transformer encoder by Kobayashi\cite{IMIB2023kobayashi} and innovative approaches like Mamba\cite{tsuji2024mamba} by Tsuji; however, they are all evaluated using only the robot’s joint information and do not handle both image and joint information like ACT\cite{act2023zhao, mact2024fu}.
To bridge these gaps, Buamanee and Kobayashi proposed Bilateral Control-Based Imitation Learning via Action Chunking with Transformers (Bi-ACT)\cite{biact2024pan}.
In Bi-ACT\cite{biact2024pan}, the effectiveness of Bi-ACT was demonstrated through a "Pick-and-Place" task involving objects of varying hardness, size, shape consistency, and weight distribution using an unimanual robot.

However, detailed analysis and validation were limited to simple tasks with an unimanual robot and adaptation to bimanual robots had not been implemented.
Therefore, this paper contributes by providing a more in-depth analysis of unimanual tasks with Bi-ACT and applying Bi-ACT to bimanual tasks by proposed ALPHA-$\alpha$" which is "A" "L"ow-cost "P"hysical "Ha"rdware Considering Diverse Motor Control Modes for Research in Everyday Bimanual Robotic Manipulation, as shown in Fig.~\ref{fig:alpha-biact}.

Our contributions are as follows:
\begin{itemize}
    \item Detailed Analysis of Bi-ACT in Unimanual Tasks: We conducted an in-depth analysis of Bi-ACT in unimanual manipulation tasks.Experimental results confirmed that Bi-ACT exhibits excellent performance and adaptability when using position and force information/control, especially in manipulating objects of different hardness, size, shape, and weight distribution.
    \item Design and Implementation of ALPHA-$\alpha$: We developed ALPHA-$\alpha$, a low-cost physical hardware considering diverse motor control modes for research in everyday bimanual robotic manipulation.
    ALPHA-$\alpha$ considers various motor control modes such as joint position, velocity, and torque, enabling researchers to create and utilize diverse control systems.
    \item Application to Bimanual Tasks Using ALPHA-$\alpha$ and Bi-ACT: We extended the application of Bi-ACT to bimanual manipulation tasks using the ALPHA-$\alpha$ via bilateral control. The combination demonstrated high success rates in coordinated bimanual operations across multiple tasks, validating the effectiveness of our approach in complex, real-world scenarios.
\end{itemize}

Our research extends from the evaluation of only unimanual arm robots in the paper\cite{biact2024pan} to provide detailed insight into the performance of Bi-ACT, introducing ALPHA-$\alpha$ and applying Bi-ACT to bimanual tasks and ALPHA-$\alpha$.
These contributions demonstrate the importance of position and force information/control in the field of robotic manipulation and the realization of a wide variety of tasks with inexpensive ALPHA-$\alpha$.
These suggest that more researchers can enter the field of bimanual robotics.

\section{Related Works}
\subsection{Data Collection Hardware for Imitation Learning}
Effective data collection is pivotal for the success of Imitation Learning (IL)\cite{lfd2009argall, lfd2022mukherjee}, often enhanced by teleoperated systems including virtual reality headsets with hand-tracking mechanisms\cite{tre2018Zhang, cheng2024opentelevision}, smartphones\cite{tre2021tung}, keyboard inputs\cite{tre2018fan}, UMI\cite{chi2024universal}, and leader-follower systems\cite{act2023zhao, mact2024fu, ge2023swu}.
Systems such as ALOHA\cite{act2023zhao}, Mobile ALOHA\cite{mact2024fu}, and GELLO\cite{ge2023swu} have significantly advanced data collection methodologies by utilizing leader-follower configurations to gather robot joint angles and image data through unilateral control.
Similarly, YAY Robot\cite{shi2024yellrobotimprovingonthefly} has developed systems that accumulate unilateral control and language information akin to the ALOHA system.
\begin{figure}[t]
  \begin{center}
    \scalebox{0.27}{
        \includegraphics{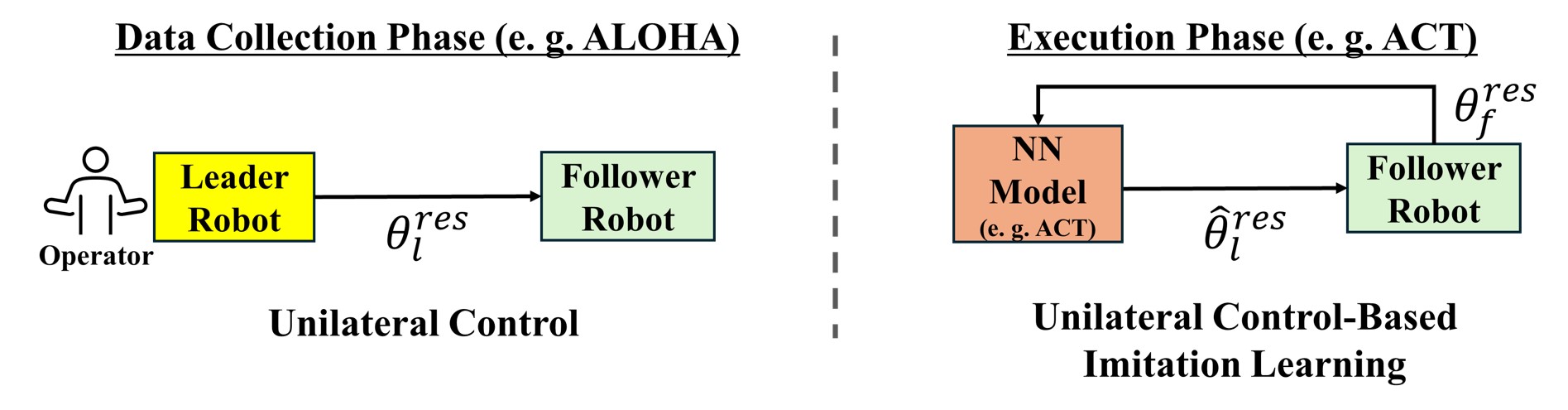}}    
  \caption{Image of Unilateral Control-Based Imitation Learning}
  \label{fig:unilateral}
\end{center}
\end{figure}

Unilateral control, wherein control commands are issued in a single direction from the operator to the robot without any sensory feedback loop to modify actions based on environmental interaction, presents certain limitations, as shown in Fig.~\ref{fig:unilateral}.
The absence of force information and feedback control complicates the accurate determination of object characteristics such as hardness and shape.
Consequently, data collection methods that employ force/tactile sensors to capture force/tactile information have been introduced\cite{liu2024forcemimic, hou2024adaptive, huang20243dvitac, kamijo2024learning}.
These methods may have limitations in the detectable frequency range of force due to the influence of sensor noise and computational speed.
Adding more sensors increases hardware complexity and costs.
Therefore, interest has increased in data collection through bilateral control systems, which facilitate the processing of both position and force information/control without necessitating force/tactile sensors\cite{biact2024pan, IMIB2022sakaino}.

This paper concentrates on data collection methods for unimanual and bimanual robotic manipulation utilizing bilateral control, which leverages both position and force information for enhanced control efficiency.
Furthermore, previous studies have highlighted the need for low-cost robots capable of accommodating diverse control inputs to the motors. In response, we developed a low-cost ALPHA-$\alpha$, which considers motor control inputs for position, velocity, and current. We also demonstrate practical applications of ALPHA-$\alpha$ via bilateral control.

\subsection{Bilateral Control-Based Imitation Learning}
\begin{figure}[t]
  \begin{center}
    \scalebox{0.26}{
        \includegraphics{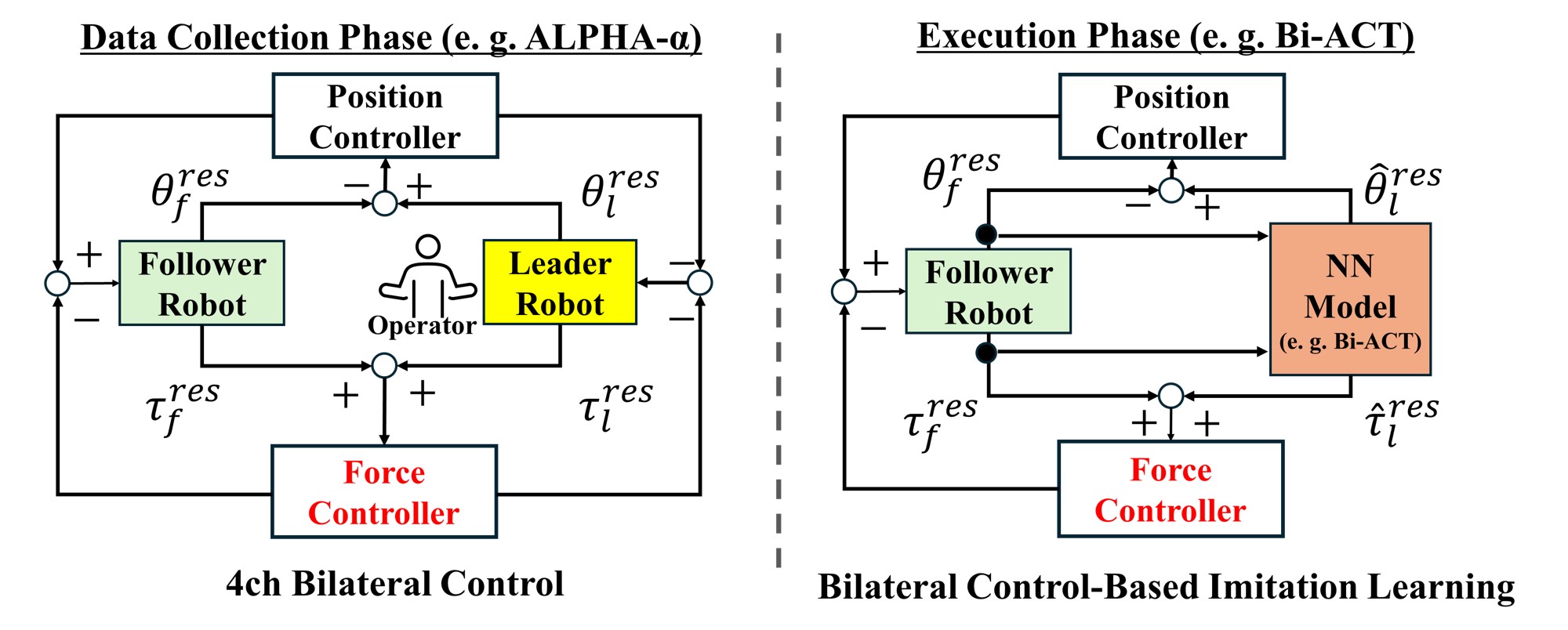}}    
  \caption{Image of 4ch Bilateral Control-Based Imitation Learning}
  \label{fig:4ch}
\end{center}
\end{figure}
Bilateral control is used for data collection which involves the remote operation of follower robots in the environment, guided by leader robots controlled by a human.
This is achieved through position tracking and the use of action-reaction principles, as shown in Fig.~\ref{fig:4ch}.
Various tasks have been accomplished using these bilateral control-based imitation learning methods. For instance, Adachi et. al. reported on using a robot to draw a line along a ruler \cite{IMIB2018adachi}. Sakaino et al. reported on slicing cucumber task \cite{IMIB2022sakaino}. These methods have previously implemented imitation learning through bilateral control using LSTM.
In addition, there are methods using transformer encoder by Kobayashi et al.\cite{IMIB2023kobayashi} and innovative approaches like Mamba\cite{tsuji2024mamba} by Tsuji.
Bilateral control enables autonomous robot operation at a speed comparable to humans, which is crucial for real-world actions.
However, most of these methods do not utilize image data for robot operation like ACT\cite{act2023zhao, mact2024fu}, limiting their adaptability to changes in the operating environment.
To bridge this gap, Buamanee and Kobayashi et al. proposed Bi-ACT\cite{biact2024pan}, a learning model based on images and robot joint positions, velocities, and torques that integrates the best aspects of bilateral control-based imitation learning and ACT.
The effectiveness of Bi-ACT was demonstrated through a "Pick-and-Place" task involving objects of varying hardness, size, shape consistency, and weight distribution using an unimanual robot. However, detailed analysis and validation were limited to simple tasks with an unimanual robot and adaptation to bimanual robots had not been implemented.

Therefore, this paper contributes by providing a more in-depth analysis of unimanual tasks and applying Bi-ACT to bimanual tasks by using ALPHA-$\alpha$ via bilateral control.
\begin{figure*}[t]
 \begin{center}
  \scalebox{1}{
  \includegraphics[width=\textwidth]{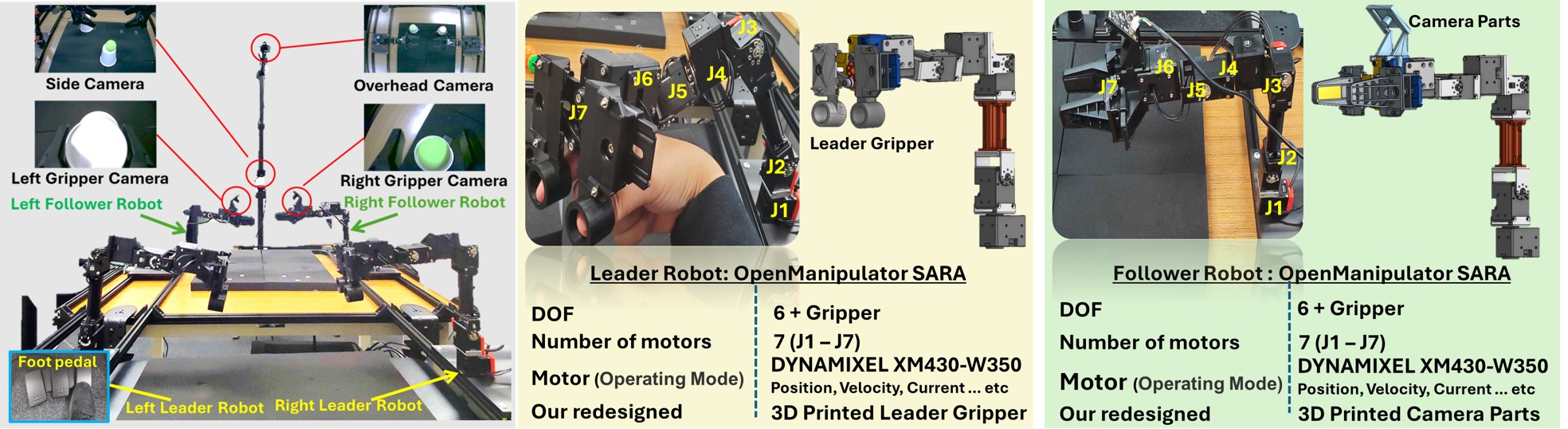}}  
 \caption{ALPHA-$\alpha$": "A" "L"ow-cost "P"hysical "Ha"rdware Considering Diverse Motor Control Modes for Research in Everyday Bimanual Robotic Manipulation}
 \label{fig:alpha}
\end{center}
\end{figure*}
\subsection{ACT: Action Chunking with Transformers}
The inception of ACT marked a leap in robotics behavior cloning algorithms\cite{act2023zhao, mact2024fu}.
ACT utilizes a Conditional Variational Autoencoder (CVAE) to model diverse scenes, along with a transformer for predicting sequences of actions (action chunking) from multimodal inputs.
This method helps mitigate errors and unpredictable responses in out-of-distribution states. 
 A multitude of models have since been developed based on ACT for imitation learning purposes, demonstrating its wide applicability and robustness\cite{act2023zhao,mact2024fu,cheng2024opentelevision, oneact2023george,Bharadhwaj_2024,kamijo2024learning,biact2024pan,chen2024exact,kareer2024egomimicscalingimitationlearning}.

Building on ACT, the One ACT Play methodology emerged as an enhancement in this domain\cite{oneact2023george}. 
One ACT Play sets itself apart by using the robot's end-effector position and posture, along with images, as inputs, in contrast to ACT's reliance on joint angles and images.
This shift enables a more intuitive and direct interaction with the robot's environment. However, neither ACT nor One ACT Play includes force information in data collection or operations.
Thus, Bi-ACT promises a more comprehensive understanding and manipulation of the robot's environment, leading to more precise and adaptable executions.
Comp-ACT proposed a Haptic Feedback Teleoperation System for rigid robots whose lowest-level actuation interface is position/velocity control\cite{kamijo2024learning}.
The key difference with these studies is that Bi-ACT collects data through bilateral control based on position and force information without using force/torque sensors, learns through Bi-ACT, and the robot operates via torque commands.

This paper provides a more detailed analysis of Bi-ACT and discusses its applicability to bimanual robot tasks by using ALPHA-$\alpha$ via bilateral control.

\section{ALPHA-$\alpha$: A Low-cost Physical Hardware Considering Diverse Motor Control Modes for Research in Everyday Bimanual Robotic Manipulation}

\subsection{Overview}
As shown in Fig.~\ref{fig:alpha-biact}-\ref{fig:alpha}, we aimed to develop ALPHA-$\alpha$, a low-cost bimanual robotic physical hardware considering diverse motor control modes that is suitable for robotics research capable of handling everyday tasks, allowing it to be easily constructed by many researchers and developers.
It is important to note that we do not claim our hardware ALPHA-$\alpha$ is superior to ALOHA in terms of performance.
The reason for comparing ALPHA-$\alpha$ and ALOHA in this paper is to clarify the position of ALPHA-$\alpha$ by comparing ALPHA-$\alpha$ with ALOHA, a bimanual robot platform used by many users.
ALPHA-$\alpha$ features low cost, ease of use, repairability, ease of assembly, and ability to enable various control types and high control frequency.

The distinctive features of ALPHA-$\alpha$ are as follows.
\begin{enumerate}
    \item \textbf{Low Cost:}
    The entire system fits within the budget of most robotics labs, costing approximately \$8,663 excluding PC equipment, which is more than half less expensive compared to the four ALOHA robots\cite{team2024aloha} at around \$19,359, as shown in Tab.~\ref{tab:cost-aloha-alpha}. 

    \item \textbf{Diverse Motor Control Modes:}
    ALPHA-$\alpha$ utilizes and improves upon robots, a robot composed of motors capable of position, velocity, and torque (current) control, to provide researchers with flexibility in selecting control methods. With the ability to control motors via position, velocity, and torque (current), a variety of control systems.

    \item \textbf{Data Collection Frequency:}
    As the selection of control systems increases, for example, sensitive manipulation by force control requires a higher control frequency. Therefore, ALPHA-$\alpha$ employs a motor capable of collecting joint angle, velocity, and torque data at 1000 Hz and an RGB camera capable of collecting RGB images at 260 Hz. For stable collection, RGB image data is collected at about 100 Hz in this paper.
\end{enumerate}
We selected and improved a robot that meets these specifications, constructing the physical hardware which we have named ALPHA-$\alpha$.
\begin{table}[t]
\caption{Cost comparison of ALOHA and ALPHA-$\alpha$}
 \scalebox{0.73}{
 \centering
\begin{tabular}{lcccc}
\hline
\textbf{System} & \textbf{4 Robots (USD)} & \textbf{4 Cameras (USD)} & \textbf{Total (USD)} & \textbf{Percentage Decrease (\%)} \\ \hline \hline
ALOHA & 19,359 & 1,126 & 20,485 & - \\
ALPHA-$\alpha$ & 8,663 & 288 & 8,951 & 56 \\
\hline 
\end{tabular}
}
\label{tab:cost-aloha-alpha}
\end{table}
\begin{table*}[t]
 \caption{Key Differences between ALOHA/ACT and ALPHA-$\alpha$/Bi-ACT}
 \centering
 \scalebox{0.74}{
\begin{tabular}{lcccc}
\hline
Approach & Control Type & Data Collection Frequency [Hz] & Model I/O Robot Dimensions & Robot Dimensions Details \\ \hline \hline
ALOHA / ACT & Unilateral control(position) & 50 & 14/14 & 14=2 (bimanual)  × 7 (6DOF+gripper) × 1 (joint angle) \\ 
ALPHA-$\alpha$ / Bi-ACT & Bilateral control(position and force) & 1000 & 42/42 & 42=2 (bimanual) × 7 (6DOF+gripper) × 3 (joint angle, velocity, torque) \\ \hline
\end{tabular}
}
\label{tab:aloha-alpha}
\end{table*}
\subsection{Design and Cost Optimization of ALPHA-$\alpha$}
The design of ALPHA-$\alpha$ was inspired by the aluminum frame structure of ALOHA\cite{team2024aloha}, but it employs a different robotic and camera system.
As of November 2024, according to the ALOHA price documentation\cite{team2024aloha}, the cost of the robots and cameras is \$20,485.96 (USD), accounting for almost 75\% of the total cost of ALOHA, which is \$27,067.41 (USD). This suggests that reducing the cost of the robots and cameras could make the hardware more accessible to a wider range of users.
ALOHA\cite{team2024aloha} utilizes four robotic arms, specifically two ViperX 300 6DOF Robot Arms and two WidowX 250 6DOF Robot Arms, with a total cost of approximately \$19,359.

In contrast, to achieve ALPHA-$\alpha$ solutions of lower cost, various motor control types, and high data collection frequency, we redesigned the OpenMANIPULATOR SARA\cite{sara} robotic arm.
OpenMANIPULATOR SARA is an example of the OpenMANIPULATOR Friends \cite{omx-friends} and is derived from OpenMANIPULATOR-X \cite{omx}.
OpenMANIPULATOR SARA is a robot composed of DYNAMIXEL XM430-W350 motors. The DYNAMIXEL XM430-W350 can operate with various control inputs such as position, velocity, and torque, and supports operation at 1000 Hz, thus meeting ALPHA-$\alpha$'s requirements for various motor control types and high data collection frequency.
As of November 2024, OpenMANIPULATOR SARA is no longer available for purchase; therefore, it is important to note that the prices listed in Tab.~\ref{tab:cost-aloha-alpha} reflect the combined cost of purchasing OpenMANIPULATOR-X and modifying it to OpenMANIPULATOR SARA. The price may change in the future.
We used four ELP-USBFHD08S-L36 cameras for ALPHA-$\alpha$. All cameras streamed at 640x360 resolution RGB images.
Thus, the total cost of robotic arms and cameras is approximately \$8,951(USD).
The purchase and assembly of ALPHA-$\alpha$ in Japan, excluding the cost of a PC, amounted to approximately 1.3 million Japanese yen(JPY).
Converting this amount into US dollars, assuming an exchange rate of 153 JPY to the \$1 USD, results in about about \$8,500. Since prices fluctuate between Japan and other countries, the robot and camera prices in Tab.~\ref{tab:cost-aloha-alpha} reflect the cost when purchased in the USA, which is higher than when purchased in Japan. However, for reference and to demonstrate significant cost reductions, we have included the price list from ALOHA.

While ALPHA-$\alpha$ is a more low-cost platform compared to ALOHA, this paper does not make any claims about the superiority or inferiority of the hardware, as the suitability of each platform depends on specific tasks and use cases.
However, an important contribution of the ALPHA-$\alpha$ hardware that we would like to emphasize is that it is inexpensive, thus enabling more researchers to build a bimanual robotic hardware system consisting of a 6-DOF arm and a gripper, which may facilitate progress in the field of robotics research.

\subsection{ALPHA-$\alpha$ via Bilateral Control}
This paper focuses on bilateral control-based imitation learning; thus, we implemented bilateral control in ALPHA-$\alpha$.
For clarity of explanation, we describe bilateral control in comparison with unilateral control, which is also employed in ALOHA.

\subsubsection{Unilateral Control}
ALOHA collects data through unilateral control, while our ALPHA-$\alpha$ adopts bilateral control. Tab.~\ref{tab:aloha-alpha} shows key differences between ALOHA and ALPHA-$\alpha$.
As shown in Fig.~\ref{fig:unilateral}, unilateral control is a method in which control commands are issued in one direction from the operator to the robot, with no sensory feedback loop to adjust actions based on interaction with the environment.
However, the lack of force information and feedback control presents a limitation, making it difficult to grasp object characteristics such as hardness and shape.

\subsubsection{Bilateral Control}
\begin{figure}[t]
  \begin{center}
    \scalebox{0.25}{
        \includegraphics{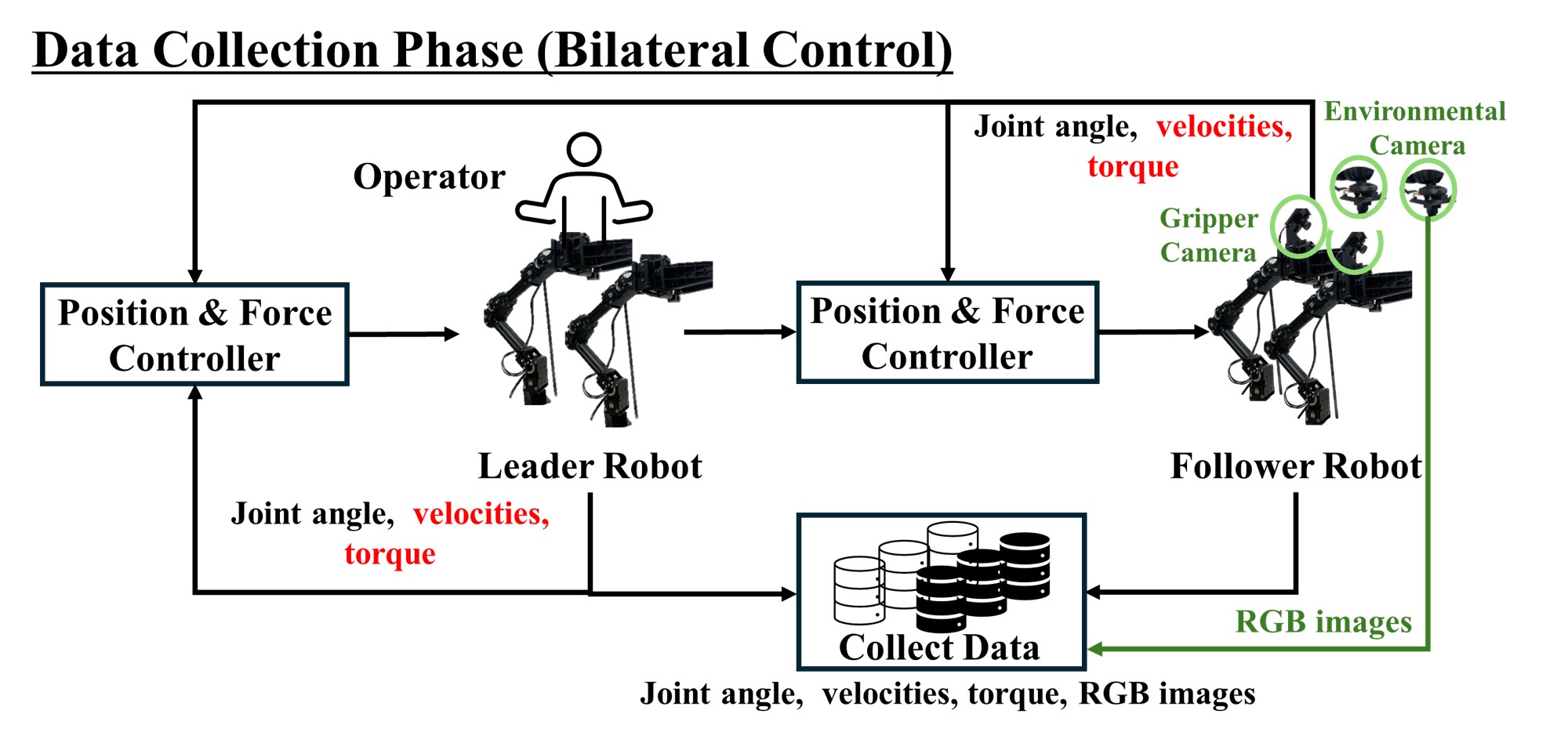}}    
  \caption{Image Diagram of Data Collection}
  \label{fig:data_bi}
\end{center}
\end{figure}

As shown in Fig.~\ref{fig:4ch}-\ref{fig:data_bi}, the controller design adopted control of position and force for each axis.
The fundamental principle of bilateral control is sharing position, force, or other information between the operator and the control target.
The control goals of the bilateral control are summarized as follows:
\begin{equation}
\theta_l - \theta_f = 0
\label{eq:position} 
\end{equation}
\begin{equation}
\tau_l + \tau_f = 0
\label{eq:force}
\end{equation}
where $\theta$ and $\tau$ represents the joint angle and torque.
The subscript $\bigcirc_l$ represents the leader system, and $\bigcirc_f$ represents the follower system.
Angle information was obtained from encoders, and angular velocity was calculated by differentiating this information. The torque response value $\tau$ was estimated using disturbance observer (DOB) \cite{DOB} and force reaction observer (RFOB) \cite{RFOB}.
This allows for the estimation of environmental reaction torques, such as force/torque, without the need for force/torque sensors, enabling the implementation of bilateral control.

Specifically, bilateral control is achieved by satisfying (\ref{eq:position}), representing position tracking between systems, and (\ref{eq:force}), representing the action-reaction relationship of forces.
As shown in Fig.~\ref{fig:4ch}, 4ch bilateral control was used to collect human operation data for IL. 
This control scheme consists of a human-operated leader and a follower robot system that tracks the leader's movements. It simultaneously controls position and torque, setting the joint angle of the leader and follower as matching target values in position control. For torque, the torque signs applied to the leader and follower joints are inverted. Thus, the follower replicates the human's operational feel, while the leader reproduces the follower's reaction force.
\subsubsection{Data Collected by ALPHA-$\alpha$ via Bilateral Control}
As shown in Fig.~\ref{fig:alpha-biact} and \ref{fig:alpha}, given that the operation of ALPHA-$\alpha$ necessitates the use of both hands, we have developed 3D-printed leader grippers. The operator manipulates the robot arm by inserting the thumb and middle finger into the holes of the operation 3D-printed leader grippers, enabling the operating robot arm and opening and closing of the grippers.
Fig.~\ref{fig:data_bi} shows image diagram of data collection of ALPHA-$\alpha$ via Bilateral Control. 
ALPHA-$\alpha$ is capable of acquiring data on joint angle, velocity, and torque for both the leader and follower at a rate of 1000 Hz. Additionally, it captures RGB image data at approximately 100 Hz using four RGB cameras. This capability stems from the fact that force/torque information is estimated without the need for direct force/torque sensors, employing bilateral control instead.
Due to the requirement that the operator uses both hands to control ALPHA-$\alpha$, the completion of data collection is facilitated by the operator stepping on a foot pedal, as depicted in Fig.~\ref{fig:alpha}.

\begin{figure*}[t]
 \begin{center}
  \scalebox{0.52}{
  \includegraphics{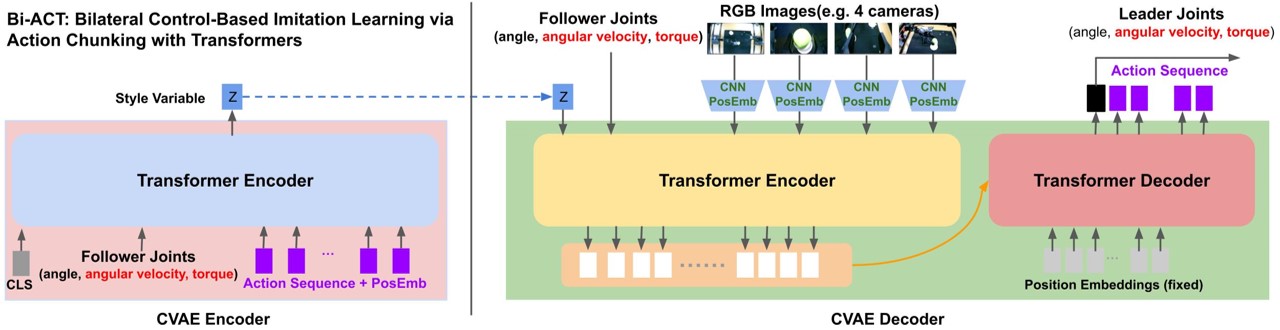}}  
 \caption{Model Architecture: Bilateral Control-Based Imitation Learning via Action Chunking with Transformers(Bi-ACT)}
 \label{fig:modelnet-biact}
\end{center}
\end{figure*}
\section{Bi-ACT: Bilateral Control-Based Imitation Learning via Action Chunking with Transformers}
\subsection{Overview}
Our proposed work employs a method inspired by ACT research, utilizing joint and image data to predict movements, combined with Bilateral Control-Based Imitation Learning principles for a robust robotic control approach, as shown in Fig.~\ref{fig:modelnet-biact}.

The primary difference between ALOHA/ACT and Bi-ACT is information and control methods. ALOHA/ACT is based on unilateral control, which relies solely on the robot's joint positions and uses the joint angle data predicted by the ACT learning model directly as command values for ALOHA's joint position control controller. This system prioritizes position targets, which can make it difficult to generate movements that require nuanced control of force. It is important to note that although it is possible to simulate force modulation in remote operations using only position control, this typically requires extensive time for operators to master the control of the leader robot.

On the other hand, our Bi-ACT is based on bilateral control, which considers the robot's joint positions, velocities, and torques. Bi-ACT utilizes not only the joint data of the leader robot—positions, velocities, and torques—but also incorporates this information from the actively operating follower robots to generate command values for current and torque control. This approach allows for control that combines both position and force in the robot's movements. Crucially, the command values are not directly generated by the model; instead, they are produced by using the values generated by the model for the leader robot in conjunction with the actual values obtained from the follower robot. By leveraging four-channel bilateral control, this method enables the generation of command values that consider interactions with the environment, thus facilitating a broader range of movements.

\begin{figure}[t]
  \begin{center}
    \scalebox{0.23}{
        \includegraphics{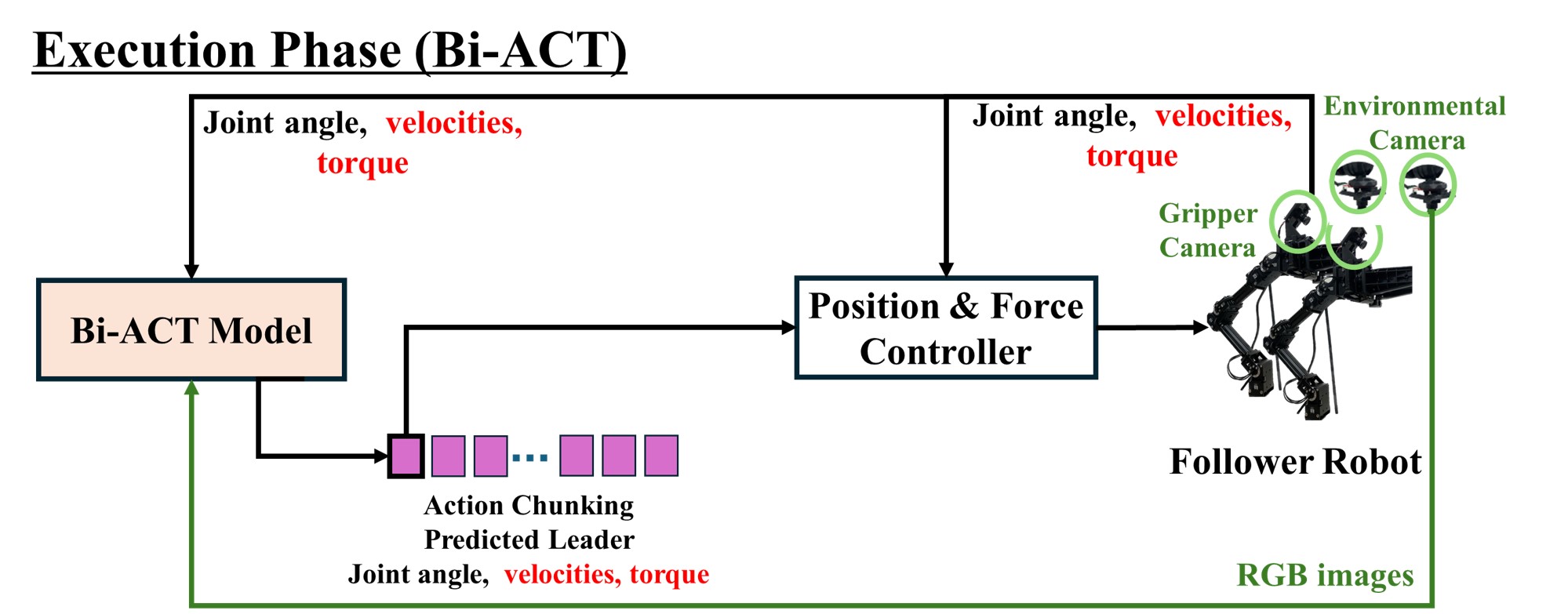}}    
  \caption{Image Diagram of Bi-ACT Execution}
  \label{fig:data_exe}
\end{center}
\end{figure}

Data collected includes images from gripper and environmental cameras, along with joint angles, angular velocities, and torque of leader and follower robots, as shown in Fig.~\ref{fig:data_bi}. Bi-ACT predicts subsequent steps for these factors, facilitating effective bilateral control in the follower robot for more responsive maneuvering, as shown in Fig.~\ref{fig:data_exe}. 
\subsection{Data Collection}

To gather demonstration data, we utilized Bilateral Control, enabling the user to operate the leader robot while sensing the environment around the follower robot, thereby enhancing data quality.
Throughout this process, we recorded joint angles, angular velocities, and torque for both robots, as well as images from the gripper and environmental cameras, as shown in Fig.~\ref{fig:data_bi}.
By incorporating force as an input, Bi-ACT accounts for differences in objects' weight and texture during training.
\subsection{Model Architecture}
To improve the environment's comprehensibility in the study, our method was based on ACT, with the addition of a novel dimension to the input and output data. Along with joint angle and image data which were previously used in the original work, we added angular velocity and torque to the input data. A schematic representation of our network structure is provided in Fig.~\ref{fig:modelnet-biact}.

Bi-ACT processes input from RGB images from the follower robot's gripper and environmental perspective. Bi-ACT also receives joint data from the follower robot, which includes joint angle, angular velocity, and torque across $N$ joints, forming $N$-dimensional vector in total. Utilizing action chunking, the policy generates $k$ x $N$ tensor for the leader robot's actions over $k$ time steps. The actions are then relayed to the controller which calculates the necessary joint currents in the follower robot for executing movements. The states of the follower and leader at the $t$-th time step are defined in (\ref{eq:follower1}) and (\ref{eq:leader1}), respectively, with the policy $\pi$ shown in (\ref{eq:policy1}).

\begin{equation}\label{eq:follower1}
f_t = [\theta_f(t), \dot\theta_f(t), \tau_f(t)]
\end{equation}
\begin{equation}\label{eq:leader1} 
l_t = [\theta_l(t), \dot\theta_l(t), \tau_l(t)]
\end{equation}
\begin{equation}\label{eq:policy1}
\pi(l_{t:t+k} | f_t)
\end{equation}
Bi-ACT structure is similar to the original ACT structure, trained as a Conditional Variational Autoencoder (CVAE), a type of generative model that creates output data using additional contextual information. Data to be input are pre-processed according to their type, as detailed below:\\\\
\textbf{Leader/Follower's Joint data}: To prevent bias, joint data, including angle, velocity, and torque in varying units (radian, radian per second, and newton per meter), are normalized before being incorporated into Bi-ACT. This normalization ensures a balanced influence on the results.\\
\textbf{Image data}: RGB images are processed using the ResNet18\cite{ResNet} backbone to extract feature maps, then flattened and enhanced with 2D sinusoidal positional embedding to preserve spatial information.

\subsection{Execution to Robot Arm}
Fig.~\ref{fig:data_exe} shows the image diagram of Bi-ACT execution.
In this paper, as the action data output by Bi-ACT in each time step consists of 3 data: joint angle, velocity, and torque of each joint, it will be transformed into the requisite current for each joint to achieve the desired state, facilitated by calculations performed by the bilateral control system. 
Given the requisite sensitivity and precision in current control, it is essential for the robot to operate at a high frequency to ensure optimal performance.
Therefore, this paper does not use Temporal Ensemble \cite{act2023zhao}.
Instead, data for the length of the action chunk is accumulated in the action chunking buffer to increase the control cycle speed, as shown in Fig.~\ref{fig:data_exe}.
Specifically, when using ALPHA-$\alpha$, the model operates at 70Hz only during inference. However, when the model does not perform inference and instead uses data from the action chunking buffer, the current command values sent to the robot are updated at 100Hz. Based on these command values, the motor operates at the same 1000Hz as data collection.

\section{Unimanual Experiments}
\begin{figure}[t]
 \begin{center}
  \scalebox{0.21}{
\includegraphics{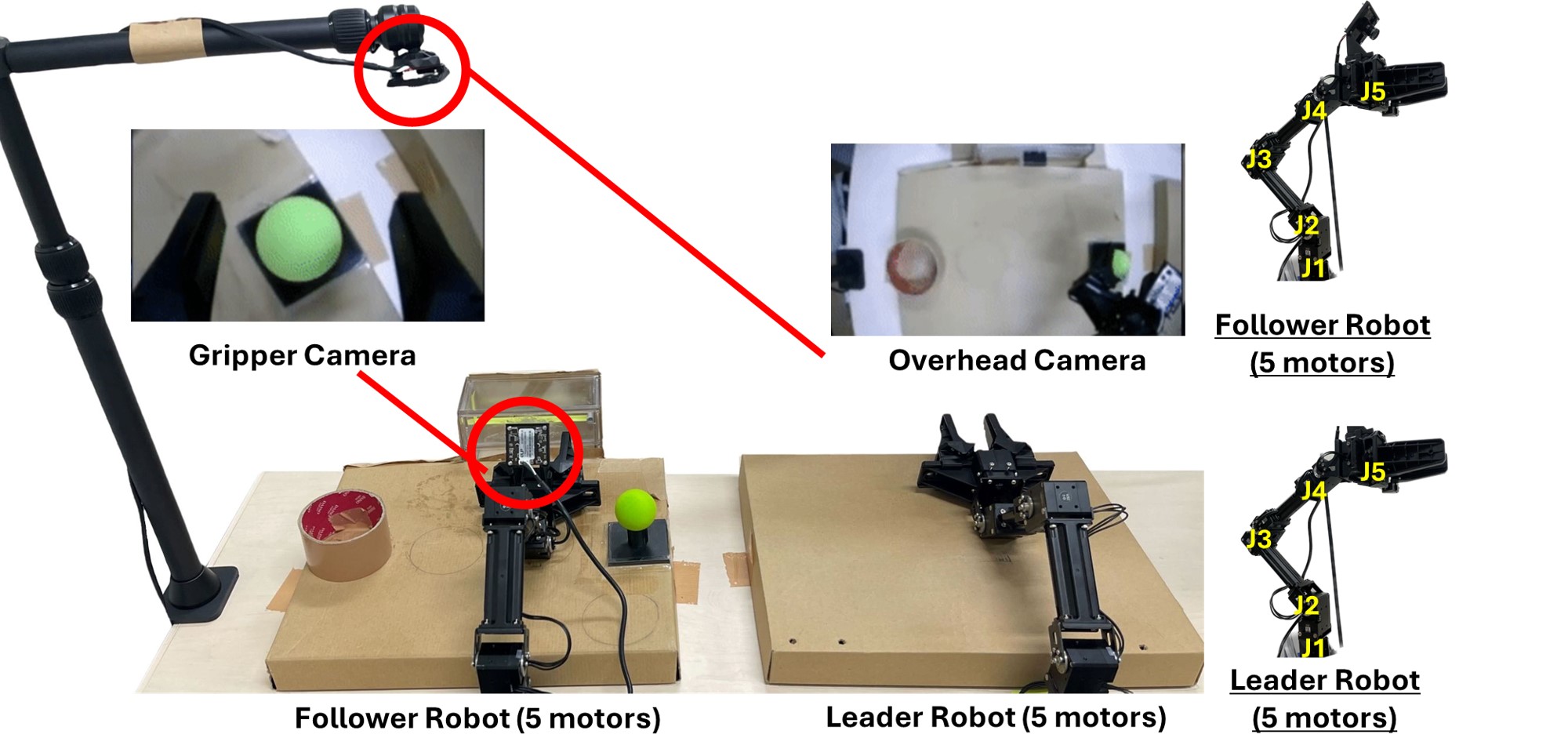}}  
 \caption{Experimental Environments Setup for Unimanual Manipulation}
 \label{fig:Environment-Setup}
\end{center}
\end{figure}
\begin{figure}[t]
 \begin{center}
  \scalebox{0.25}{
  \includegraphics{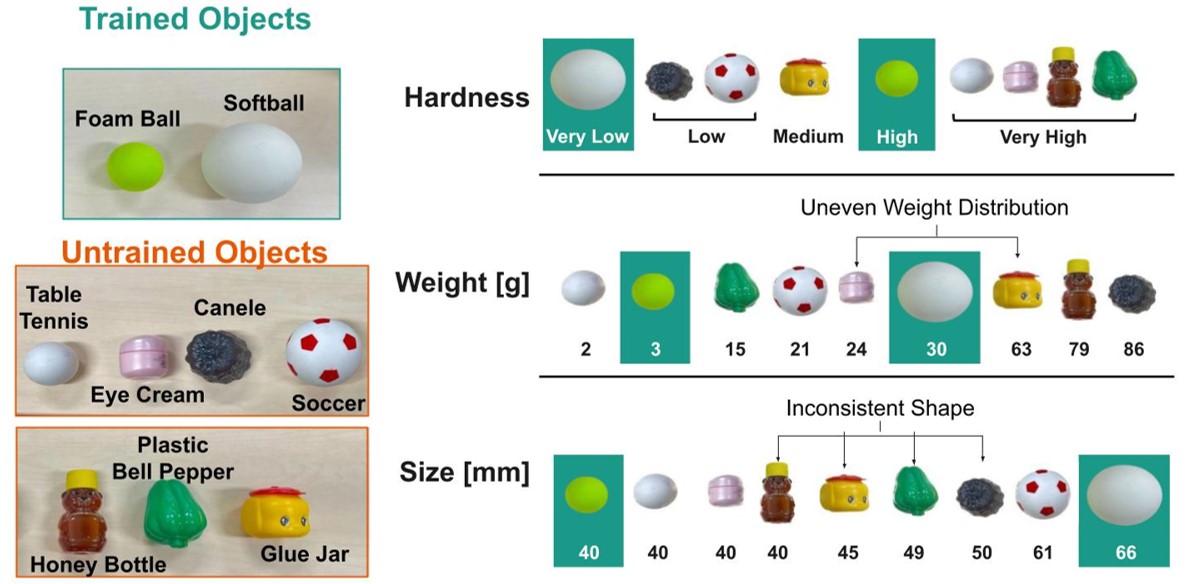}}  
 \caption{Trained and Untrained Objects in Unimanual Experiment}
 \label{fig:Target-Object}
\end{center}
\end{figure}

\begin{table}[t]
 \caption{Detail of Trained and Untrained Objects in Unimanual Experiment}
  \scalebox{0.85}{
\begin{tabular}{ccccc}
\hline
 & Training Data & Size [mm] & Hardness & Weight [g]\\ \hline \hline
Foam ball & \ding{52} & 40 & High & 3 \\ 
softball & \ding{52} & 66 & Very Low & 30 \\ 
Table tennis & \ding{54} & 40 & Very High & 2 \\ 
Eye cream  & \ding{54} & 40 & Very High & 24 \\ 
Canele & \ding{54} & 50 & Low & 86 \\ 
Soccer & \ding{54}& 61 & Low & 21 \\ 
Honey bottle & \ding{54} & 40 & Very High & 79 \\ 
Plastic Bell Pepper & \ding{54}& 49 & Very High & 15 \\ 
Glue Jar & \ding{54}& 45 & Medium & 63 \\ \hline
\end{tabular}
}
\label{tab:data_obj}
\end{table}
\begin{table}[t]
    \centering
    \caption{Unimanual Experimental Results: Pick-and-Place}
    \scalebox{0.75}{
    \begin{tabular}{c|cc|c|c|c|c}
        \hline
        \multicolumn{1}{c}{\multirow{2}{*}{Model}} & \multicolumn{2}{c}{\multirow{2}{*}{Objects}} &\multicolumn{4}{c}{Pick-and-Place}\\
        \multicolumn{1}{c}{}&\multicolumn{1}{c}{}&\multicolumn{1}{c}{}&\multicolumn{1}{c}{Pick\#1}&\multicolumn{1}{c}{Move\#2}&\multicolumn{1}{c}{ Place\#3}&Total\\\hline \hline
        \multirow{8}{*}{LSTM} &\multirow{2}{*}{Trained}&Softball&100&100&100&100\\
        \multirow{8}{*}{} &&Foam ball&70&70&70&70\\\cline{2-7}
        &\multirow{7}{*}{Untrained}&Table Tennis&90&90&90&90\\
        &&Eye Cream&70&40&40&40\\
        &&Canele&70&40&40&40\\
        &&Soccer Ball&70&70&60&60\\\
        &&Honey Bottle&70&60&60&60\\\
        &&Plastic Bell Pepper&0&0&0&0\\
        &&Glue Jar&50&50&50&50\\\hline
        \multirow{6}{*}{Bi-ACT} &\multirow{2}{*}{Trained}&Softball&80&80&80&80\\
        \multirow{6}{*}{(w/o Force)} &&Foam ball&100&100&100&100\\\cline{2-7}
        &\multirow{3}{*}{Untrained}&Table Tennis&100&100&100&100\\
        &&Eye Cream&70&50&50&50\\
        &&Canele&100&80&80&80\\
        &&Soccer Ball&90&90&80&80\\
        &&Honey Bottle&90&90&90&90\\\
        &&Plastic Bell Pepper&70&70&70&70\\
        &&Glue Jar&50&50&50&50\\\hline
        \multirow{8}{*}{Bi-ACT} &\multirow{2}{*}{Trained}&Softball&100&100&100&100\\
        \multirow{8}{*}{(w/ Force)} &&Foam ball&100&100&100&100\\\cline{2-7}
        &\multirow{7}{*}{Untrained}&Table Tennis&100&100&100&100\\
        &&Eye Cream&100&100&100&100\\
        &&Canele&100&80&80&80\\
        &&Soccer Ball&90&90&90&90\\\
        &&Honey Bottle&90&90&90&90\\\
        &&Plastic Bell Pepper&80&80&80&80\\
        &&Glue Jar&80&80&80&80\\\hline
    \end{tabular}
    }
    \label{tab:Experimental_Results-Pick}
\end{table}
\begin{figure*}[t]
 \begin{center}
  \scalebox{0.5}{
  \includegraphics{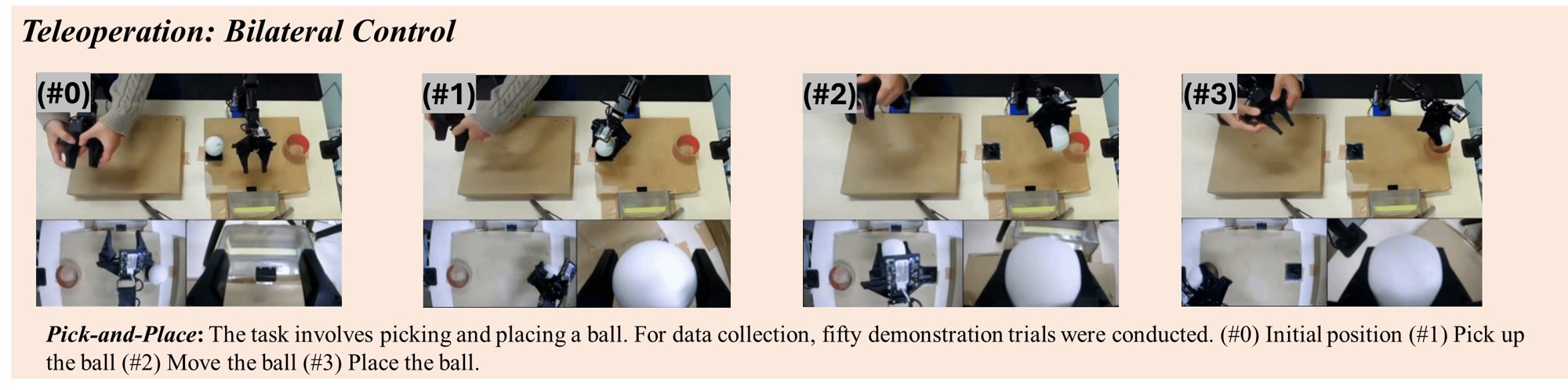}}  
 \caption{Data Collection of Unimanual Experiment}
 \label{fig:data_colection}
\end{center}
\end{figure*}

In this unimanual experiment, we compared Bi-ACT with other methods and models that do not include force control, to conduct a fundamental evaluation of Bi-ACT.
Specifically, we aimed to analyze how positional and force information/control affect the efficacy of imitation learning in robotic arms.

\subsection{Hardware}
As shown in Fig.~\ref{fig:Environment-Setup}, the OpenMANIPULATOR-X robotic arms developed by ROBOTIS were used for the experiments of unimanual robotic manipulation. A total of two robots were utilized in the experiments, including one leader robot operated by the human operator and one follower robot. Each robot has 4 degrees of freedom (DOF) for versatile movement, as well as an additional DOF for the gripper, utilizing a total of 5 motors for its operation.
The control cycle was set to 1000 Hz for precise movement. Furthermore, two RGB cameras were positioned overhead and in the gripper area of the follower robot to record observations.
\subsection{Task Setting}
As shown in Fig.~\ref{fig:Environment-Setup}-\ref{fig:data_colection}, the 'Pick-and-Place' task aimed to determine the accuracy of grippers in handling objects of various shapes, weights, and textures from the pick area and transporting them to the place area. The leader and follower robots were positioned adjacent to each other. The task execution area was set up on the follower robot's side.
During data collection, the foam ball and softball were used. For model testing, these two objects, along with seven untrained objects - the table tennis ball, an eye-cream package, Canele, the soccer ball, the plastic bell pepper, the honey bottle, and the glue jar - were tested as shown in Fig.~\ref{fig:Target-Object} and Tab.~\ref{tab:data_obj}. The success requirement was for the grippers to transport these objects without dropping them during transit; any dropping outside the designated place area was considered a failure.

\subsection{Training Dataset}
A total of 50 demonstration episodes were captured, comprising 25 trials using a foam ball and 25 trials using a softball.
Joint angles, angular velocities, and torque data were collected from the Leader and Follower robots using a bilateral control system. The robot was controlled at a frequency of 1000Hz. We adjusted the data to 100Hz for training datasets.
Based on the dataset, we trained LSTM, Bi-ACT without force, and Bi-ACT with force.

\subsection{Experimental Results}
\begin{figure*}[t]
 \begin{center}
  \scalebox{0.6}{
  \includegraphics{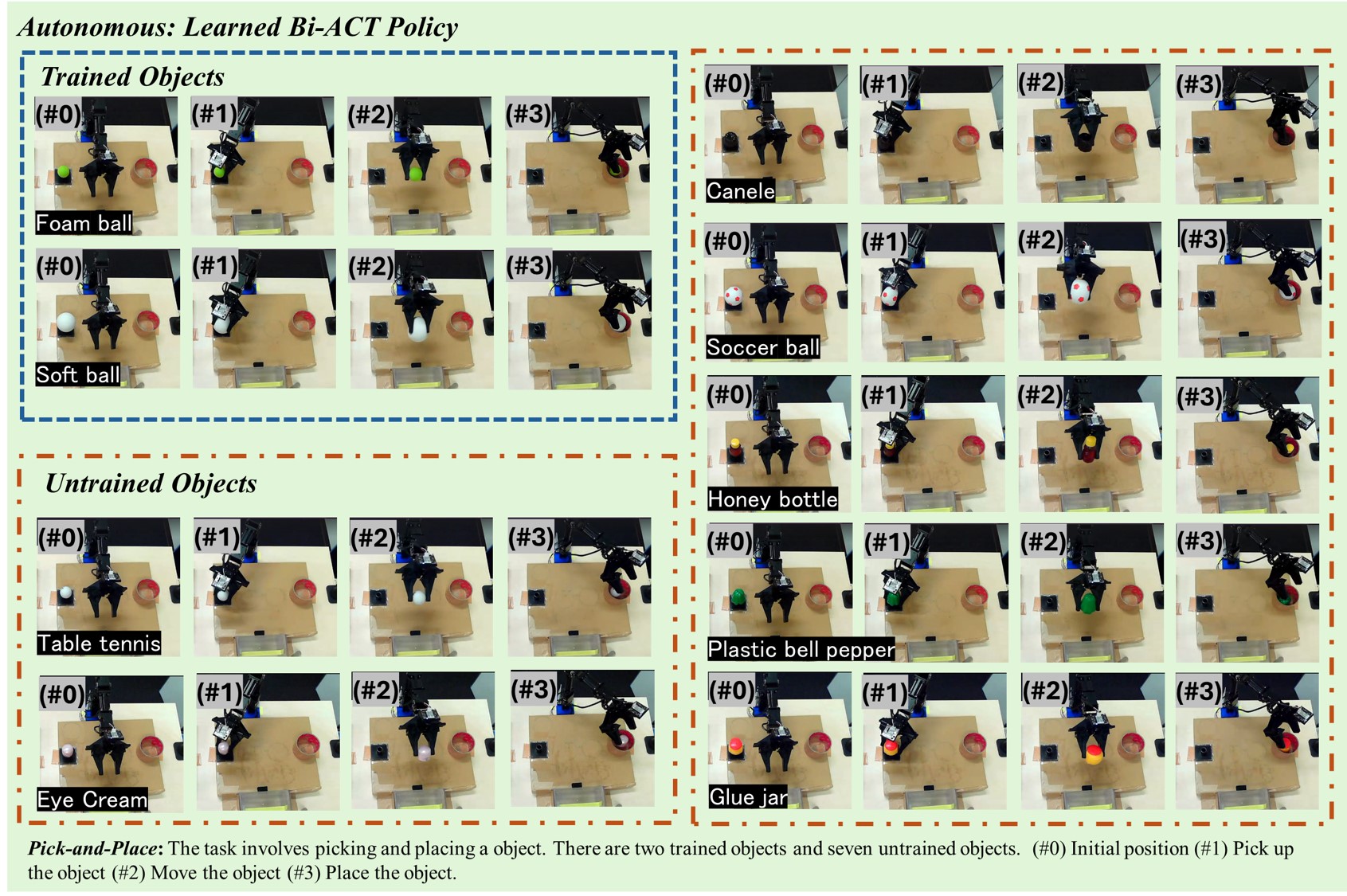}}  
 \caption{Experimental Results of Unimanual Experiment}
 \label{fig:data_results}
\end{center}
\end{figure*}
Performance was assessed by comparing LSTM, Bi-ACT without force control(with only position control), and Bi-ACT with position and force control.
Evaluations were conducted on two trained and seven untrained objects, as shown in Fig.~\ref{fig:data_results}.
Tab.~\ref{tab:Experimental_Results-Pick} shows the model's percentage accuracy across 10 trials for each object. 

LSTM performed adequately on some trained objects, achieving a 100\% success rate with the softball. However, it struggled with untrained objects, particularly those with varying consistencies and inconsistent shapes. For instance, the model’s success rate dropped to 40\% for eye cream container and was unable to handle complex shapes like the plastic bell pepper, which it failed to manipulate entirely.

Bi-ACT without force control operated effectively with smaller objects, such as foam balls and table tennis balls, but its effectiveness diminished when dealing with larger or deformable objects or those with inconsistent shapes. This variation in performance accentuates the vital role of force feedback in augmenting the model's adaptability to objects with intricate geometries and varying consistencies.

Bi-ACT with force control achieved a 100\% success rate in handling balls of various sizes and shapes, including both trained and two untrained objects.
It also demonstrated high accuracy with five other untrained objects: canele, soccer ball, honey bottle, plastic bell pepper, and glue jar. This performance indicates substantial effectiveness despite the model lacking prior training on these items.
The most significant difference between Bi-ACT model and the one lacking force data was in handling eye cream and glue jar; the latter model demonstrated notably lower performance. Both items, containing liquid, possess inconsistent weight distribution which likely posed significant challenges for the model without force data, highlighting the importance of force data in managing such objects.
\subsection{Analysis of Bi-ACT}
In this section, we analyze Bi-ACT model's performance on various objects, focusing on joint5(OpenMANIPULATOR-X)—the gripper joint with the most contact with the objects.
Results showed that integrating force metrics significantly enhanced its effectiveness. Details are as follows.
\begin{figure}[t]
 \begin{center}
  \scalebox{0.38}{
  \includegraphics{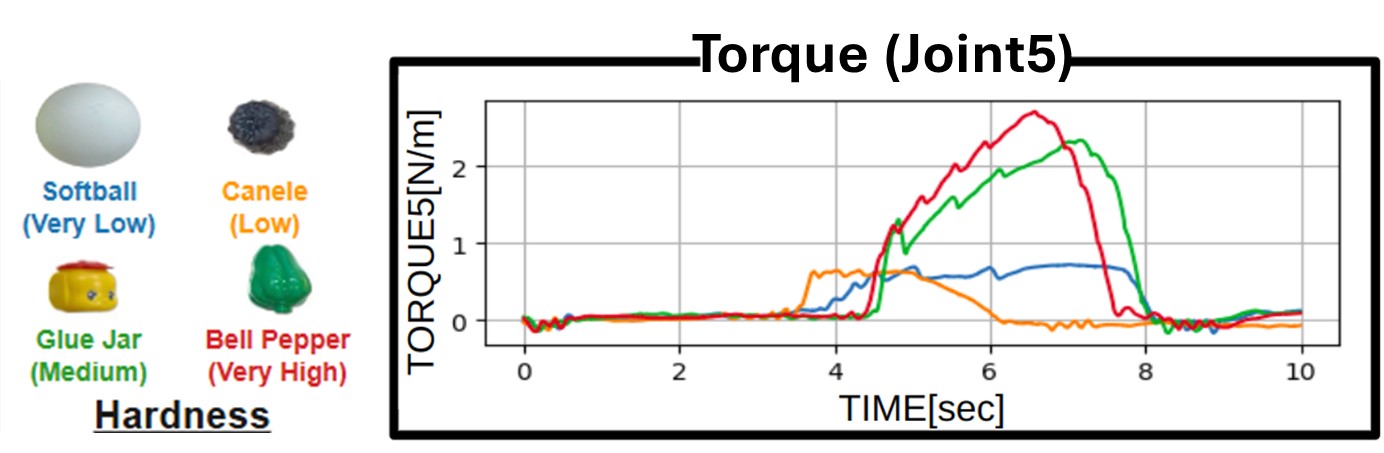}}  
 \caption{Effect of Hardness (Bi-ACT)}
 \label{fig:analyze-hardness}
\end{center}
\end{figure}

\subsubsection{Difference in hardness}
As shown in Fig. \ref{fig:analyze-hardness}, we observe the impact of object hardness on torque exerted at joint5. The torque data from the follower shows a clear trend: harder objects require more gripping force. Among the tested objects, the plastic bell pepper—with its slippery surface or unique shape—demanded the highest force, followed by the glue jar. This indicates that objects with either a high hardness level or challenging surface characteristics, like slipperiness or inconsistent shape, require increased force for stable manipulation. On the other hand, softer objects like the canele and softball showed lower force levels, with the softball, in particular, demonstrating a gradual force increase due to its larger size and lower hardness. These observations emphasize the model's ability to adapt to different hardness levels and indicate that force control allows the robot to apply appropriate gripping power according to each object’s physical properties.

\subsubsection{Diﬀerence in shape consistency}
\begin{figure}[t]
 \begin{center}
  \scalebox{0.36}{
  \includegraphics{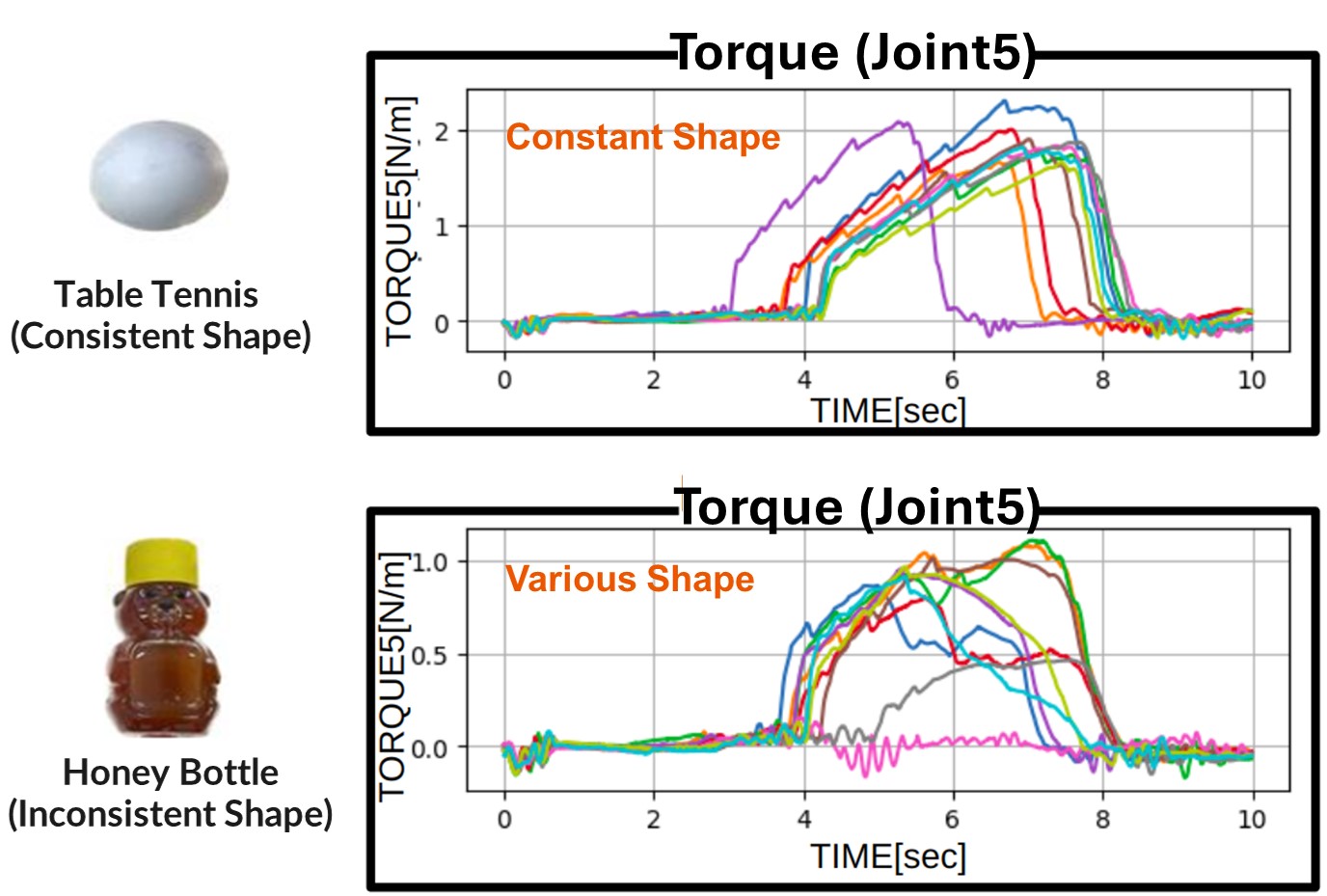}}  
 \caption{Effect of Shape Consistency - 10 trials (Bi-ACT)}
 \label{fig:analyze-shapeconsistency}
\end{center}
\end{figure}

As shown in Fig. \ref{fig:analyze-shapeconsistency}, we analyzed the impact of shape consistency by comparing two objects: the table tennis ball (consistent shape) and the honey bottle (inconsistent shape). The torque readings for the table tennis ball remained uniform across the 10 trials, as its consistent shape provided predictable points of contact for the gripper. However, the torque values for the honey bottle varied significantly, likely due to its inconsistent shape, which changes the contact points between the gripper and the object with each attempt. This variation underscores the model's responsiveness to inconsistent shapes and highlights the importance of force feedback in adapting to inconsistent contact points during manipulation.

These results confirmed the importance of position and force information/control when using Bi-ACT.
\begin{figure*}[t]
 \begin{center}
  \scalebox{0.59}{
  \includegraphics{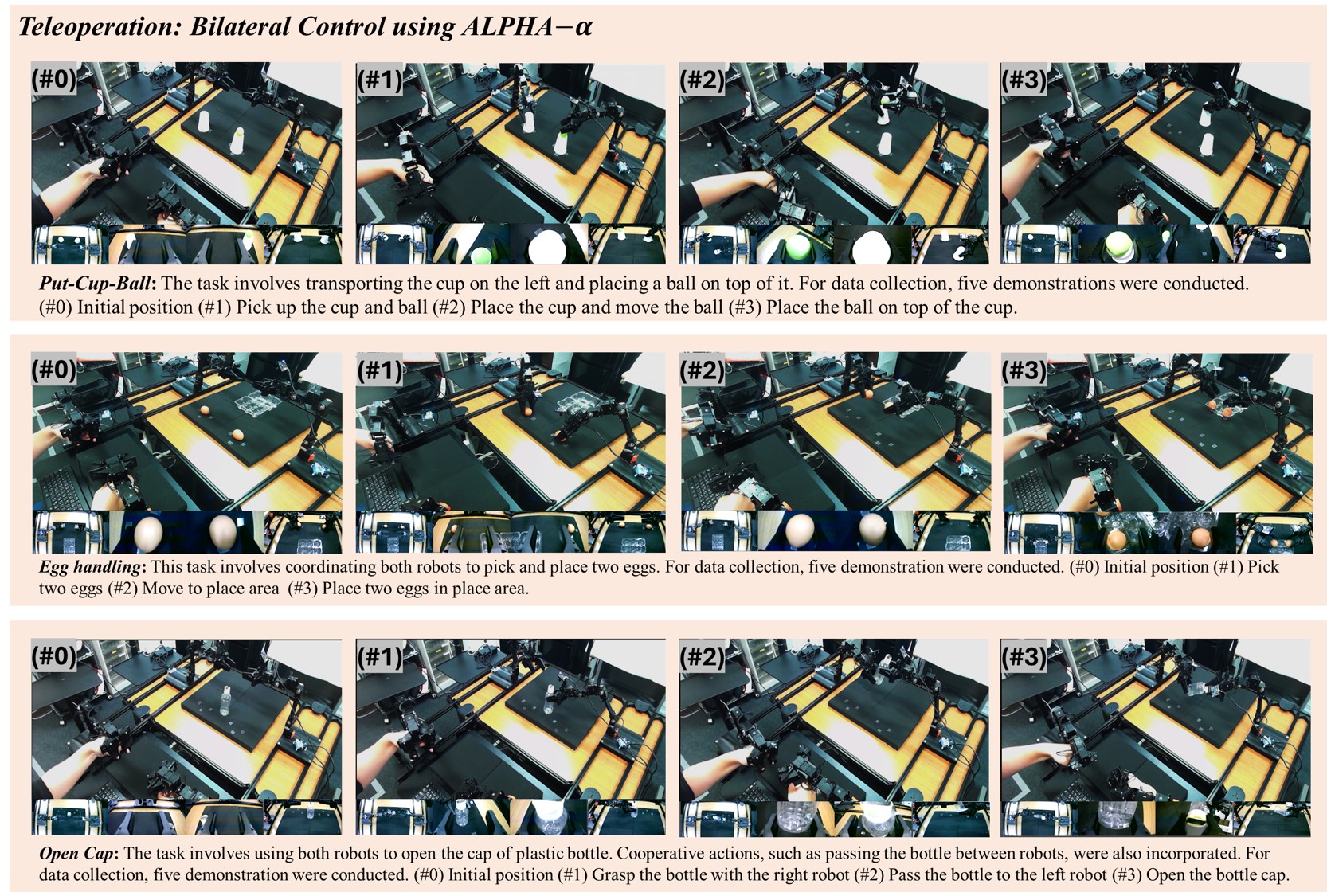}}  
 \caption{Data Collection of Bimanual Experiments by using ALPHA-$\alpha$ via Bilateral Control}
 \label{fig:bi-task}
\end{center}
\end{figure*}
\section{Bimanual Experiments}
In the bimanual experiments, we aimed to verify the applicability of the ALPHA-$\alpha$ and Bi-ACT system for complex bimanual tasks.
Specifically, we conducted tasks such as "Placing a Ball in a Cup," "Egg Handling," and "Cap Opening" to assess the performance of the Bi-ACT and the efficiency of coordinated bimanual actions in these scenarios.
\subsection{Hardware}
\begin{figure*}[t]
 \begin{center}
  \scalebox{0.53}{
  \includegraphics{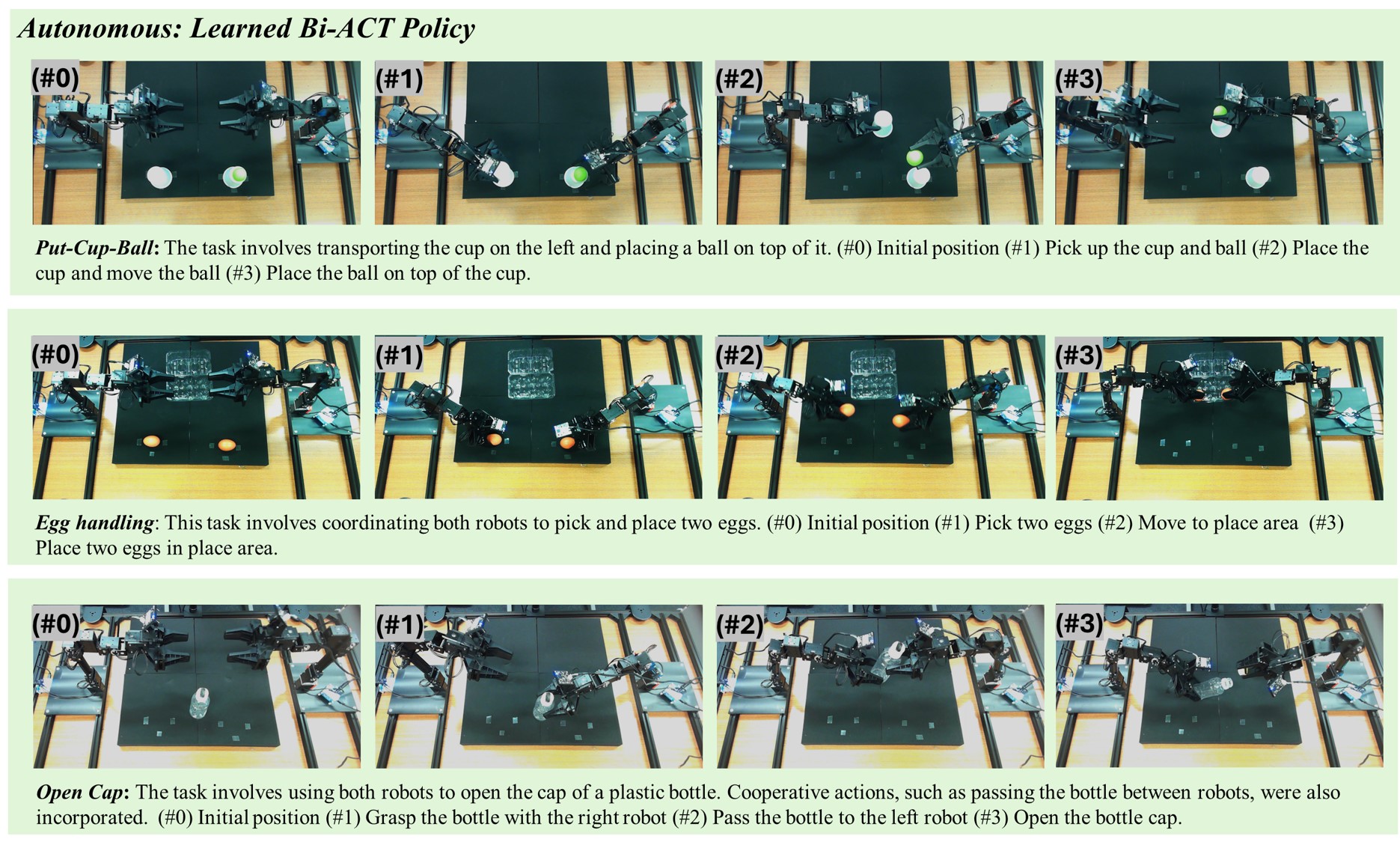}}  
 \caption{ Results of Bimanual Experiments by using ALPHA-$\alpha$}
 \label{fig:alpha-result}
\end{center}
\end{figure*}
As shown in Fig.~\ref{fig:alpha}, ALPHA-$\alpha$ were used for experiments of bimanual robotic manipulation.
A total of four robots were utilized in the experiments, including two leader robots operated by the human operator and two follower robots. Each robot has 6 degrees of freedom (DOF) for versatile movement, as well as an additional DOF for the gripper, utilizing a total of 7 motors for its operation.
The bilateral control cycle was set to 1000 Hz for precise movement and data collection of joint angle, velocity, and torque.
Furthermore, four RGB cameras were placed overhead, on the sides, and at both the right and left gripper areas of the follower robots to record observations.

\subsection{Task Setting}
To examine the applicability of Bi-ACT, experiments were conducted on three tasks, "Put-Cup-Ball," "Egg Handling," and "Open Cap," using ALPHA-$\alpha$ as shown in Fig.~\ref{fig:bi-task}. For each task, we collected 5 demonstrations as training data.

\subsubsection{Put-Cup-Ball}
In the "Put-Cup-Ball" task, the left robot arm transports a cup located on the left side, while the right robot arm picks up a ball and places it on top of the cup. The specific steps are as follows: (\#0) Initial position (\#1) Pick up the cup and ball (\#2) Place the cup and move the ball (\#3) Place the ball on top of the cup.

This task requires coordinated bimanual robot actions, as each arm must monitor the other's status and avoid interference, ensuring effective cooperation between the two robot arms.

\subsubsection{Egg Handling}
In the "Egg Handling" task, the two robots coordinate to lift two eggs and place them in a designated area. The specific steps are as follows:  (\#0) Initial position (\#1) Pick two eggs (\#2) Move to place area  (\#3) Place two eggs in place area.

This task requires the left and right arms to carefully grasp and transport fragile eggs to the specified location. Proper and delicate bimanual robot actions are essential to avoid breaking the eggs.

\subsubsection{Open Cap}
In the "Open Cap" task, two robots are used to open the cap of a plastic bottle. The specific steps are as follows: (\#0) Initial position (\#1)  Grasp the bottle with the right robot (\#2) Pass the bottle to the left robot (\#3) Open the bottle cap.

This task requires even more careful coordination than the "Put-Cup-Ball" task, as both arms must monitor each other’s status and avoid interference. In particular, it necessitates coordinated bimanual actions, such as passing the bottle between robots and holding the bottle with the left robot arm while the right robot arm opens the cap.

\subsection{Training Dataset}
We utilized the Data Augmentation Method for Bilateral Control-Based Imitation Learning with Images (DABI)\cite{kobayashi2024dabi} as the training data for the Bi-ACT model.
DABI can be employed for data augmentation when high-frequency robot data and lower-frequency image data are collected.
In this case, we first collected five demonstration data for each of the tasks: "Put the Cup," "Handle Eggs," and "Open the Cap," to serve as the training data for the Bi-ACT model.
By applying DABI, we expanded the training dataset from 5 demonstrations to 50 demonstrations. This was achieved by downsampling the data collected at 1000 Hz to 100 Hz, effectively increasing the dataset size by a factor of 10.
For further details, please refer to DABI\cite{kobayashi2024dabi}.

\subsection{Experimental Results}
\begin{table}[t]
    \centering
    \caption{Success Rate of Bimanual Experiments}
    \scalebox{0.95}{
    \begin{tabular}{c c c c c c}
        \hline
        \multicolumn{1}{c}{\multirow{1}{*}{Method}} & \multicolumn{1}{c}{\multirow{1}{*}{Task}}&\multicolumn{1}{c}{\#1}&\multicolumn{1}{c}{\#2}&\multicolumn{1}{c}{\#3}&\multicolumn{1}{c}{Total}\\\hline \hline
        \multirow{2}{*}{ALPHA-$\alpha$/Bi-ACT}
        &Put-Cup-Ball&100&100&100&100\\
        &Egg Handling&80&80&80&80\\
        &Open Cap&100&100&80&80\\  \hline
    \end{tabular}
    }
    \label{tab:alpha-task}
\end{table}
As shown in Fig.~\ref{fig:alpha-result} and Tab.~\ref{tab:alpha-task}, the performance of Bi-ACT model, incorporating force control, was evaluated to assess its applicability to bimanual tasks using the ALPHA-$\alpha$. The results of five trials for each task are presented in Tab.~\ref{tab:alpha-task}. The findings demonstrated that Bi-ACT model with force control, utilizing the ALPHA-$\alpha$, exhibited high success rates and confirmed its applicability to bimanual tasks.

\subsubsection{Put-Cup-Ball} As shown in Fig.~\ref{fig:alpha-result} and Tab.~\ref{tab:alpha-task}, the task success rate was 100\%. 
This task required the bimanual robot to perform coordinated actions and avoid mutual interference, showcasing successful task execution that cannot be fully assessed using unimanual robot arm tasks.

\subsubsection{Egg Handling} As shown in Fig.~\ref{fig:alpha-result} and Tab.~\ref{tab:alpha-task}, the task success rate was 80\%. In this task, the left and right robot arms needed to carefully grasp and transport fragile eggs to the specified location. Proper and delicate bimanual robot actions were essential to avoid breaking the eggs.
Although no eggs were broken during either data collection or autonomous operation, there was one instance in which the robot failed to grasp an egg. The egg was not damaged, but due to its round shape and slippery surface, the robot missed the grasping attempt.

\subsubsection{Open Cap}
As shown in Fig.~\ref{fig:alpha-result} and Tab.~\ref{tab:alpha-task},  the task success rate was 80\%, with one failure in the cap-opening action. This task required even higher levels of coordination and interference management between the two arms compared to the Put-Cup-Ball task, enabling the evaluation of task performance that could not be fully assessed with the Put-Cup-Ball task. Specifically, in step \#1, the right robot arm grasped the bottle; in step \#2, the left robot arm received the bottle from the right arm; and in step \#3, the left robot arm held the bottle while the right robot arm rotated the cap to open it. This task was highly contact-intensive and required significant bimanual coordination.

These results indicated that Bi-ACT model, based on bilateral control utilizing position and force information, is applicable for executing bimanual tasks.

\section{Conclusion}
This paper introduced ALPHA-$\alpha$, a low-cost physical hardware considering diverse motor control modes for research in everyday bimanual robotic manipulation, and Bi-ACT for unimanual and bimanual robotic manipulation. We demonstrated the importance of position and force information/control and showed the effectiveness of Bi-ACT and ALPHA-$\alpha$ through comprehensive real-world experiments.

Our experiments confirmed that Bi-ACT exhibits excellent performance and adaptability when using position and force information/control, especially in manipulating objects of different hardness, size, shape, and weight distribution. The application of Bi-ACT to bimanual tasks using ALPHA-$\alpha$ via bilateral control demonstrated high success rates in coordinated bimanual operations across multiple tasks.

These experiments for both unimanual and bimanual manipulation demonstrated Bi-ACT's potential, confirmed the importance of position and force information/control, and showed the effectiveness of Bi-ACT and ALPHA-$\alpha$.
However, there are certain limitations. The limitations and future challenges are as follows:
\begin{itemize}
    \item Expanded Evaluations: We plan to conduct evaluations with a wider range of robots and tasks to further validate our approach.
    \item Hardware Structure of ALPHA-$\alpha$: While ALPHA-$\alpha$ is inexpensive and allows researchers to build bimanual robotic hardware.
    However, ALPHA-$\alpha$ has differences in hardware structure compared to ALOHA, such as the length of the robotic arms.
\end{itemize}

\bibliographystyle{IEEEtran}


\end{document}